\pdfoutput=1
\documentclass{article} 
\PassOptionsToPackage{table}{xcolor}
\usepackage[final]{colm2025_conference}

\usepackage{microtype}
\usepackage{hyperref}
\usepackage{url}
\usepackage{booktabs}
\usepackage{arydshln}
\usepackage{lineno}

\definecolor{darkblue}{rgb}{0, 0, 0.5}
\hypersetup{colorlinks=true, citecolor=darkblue, linkcolor=darkblue, urlcolor=darkblue}

\usepackage{multirow}
\usepackage{graphicx}

\usepackage{amsmath}

\usepackage{makecell} 
\usepackage{caption} 

\usepackage[most]{tcolorbox}
\usepackage{float}
\usepackage{xspace}
\tcbset{
  aibox/.style={
    width=\textwidth,
    top=10pt,
    colback=white,
    colframe=black,
    colbacktitle=black,
    enhanced,
    center,
    attach boxed title to top left={yshift=-0.1in,xshift=0.15in},
    boxed title style={boxrule=0pt,colframe=white,},
  }
}
\newtcolorbox{AIbox}[2][]{aibox,title=#2,#1}
\definecolor{deepgreen}{rgb}{0.0, 0.5, 0.0}

\usepackage{listings}

\title{ShareGPT-4o-Image: Aligning Multimodal Models with \\GPT-4o-Level Image Generation}

\author{Junying Chen$^\dagger$, Zhenyang Cai$^\dagger$, Pengcheng Chen$^\dagger$, Shunian Chen$^\dagger$,\\
\textbf{Ke Ji, Xidong Wang, Yunjin Yang, Benyou Wang$^*$}\\
The Chinese University of Hong Kong, Shenzhen\\
\textit{wangbenyou@cuhk.edu.cn}\\
{{\url{https://github.com/FreedomIntelligence/ShareGPT-4o-Image}}}
}

\begin{document}

\ifcolmsubmission
\linenumbers
\fi

\maketitle

\renewcommand{\thefootnote}{\fnsymbol{footnote}}
\footnotetext[2]{Equal Contribution. $^*$Corresponding author.}
\renewcommand{\thefootnote}{\arabic{footnote}}

\vspace{-2mm}
\begin{abstract}
\vspace{-2mm}
Recent advances in multimodal generative models have unlocked photorealistic, instruction-aligned image generation, yet leading systems like GPT-4o-Image remain proprietary and inaccessible. To democratize these capabilities, we present \textbf{ShareGPT-4o-Image}, the first dataset comprising 45K text-to-image and 46K text-and-image-to-image data, all synthesized using GPT-4o's image generation capabilities for distilling its advanced image generation abilities. Leveraging this dataset, we develop \textbf{Janus-4o}, a multimodal large language model capable of both text-to-image and text-and-image-to-image generation. Janus-4o not only significantly improves text-to-image generation over its predecessor, Janus-Pro, but also newly supports text-and-image-to-image generation. Notably, it achieves impressive performance in text-and-image-to-image generation from scratch, using only 91K synthetic samples and 6 hours of training on an 8×A800 GPU machine. We hope the release of ShareGPT-4o-Image and Janus-4o will foster open research in photorealistic, instruction-aligned image generation.
\end{abstract}

\begin{figure*}[ht!]
\vspace{-1mm}
  \centering
  \resizebox{0.86\textwidth}{!}{
  \includegraphics[width=\textwidth]{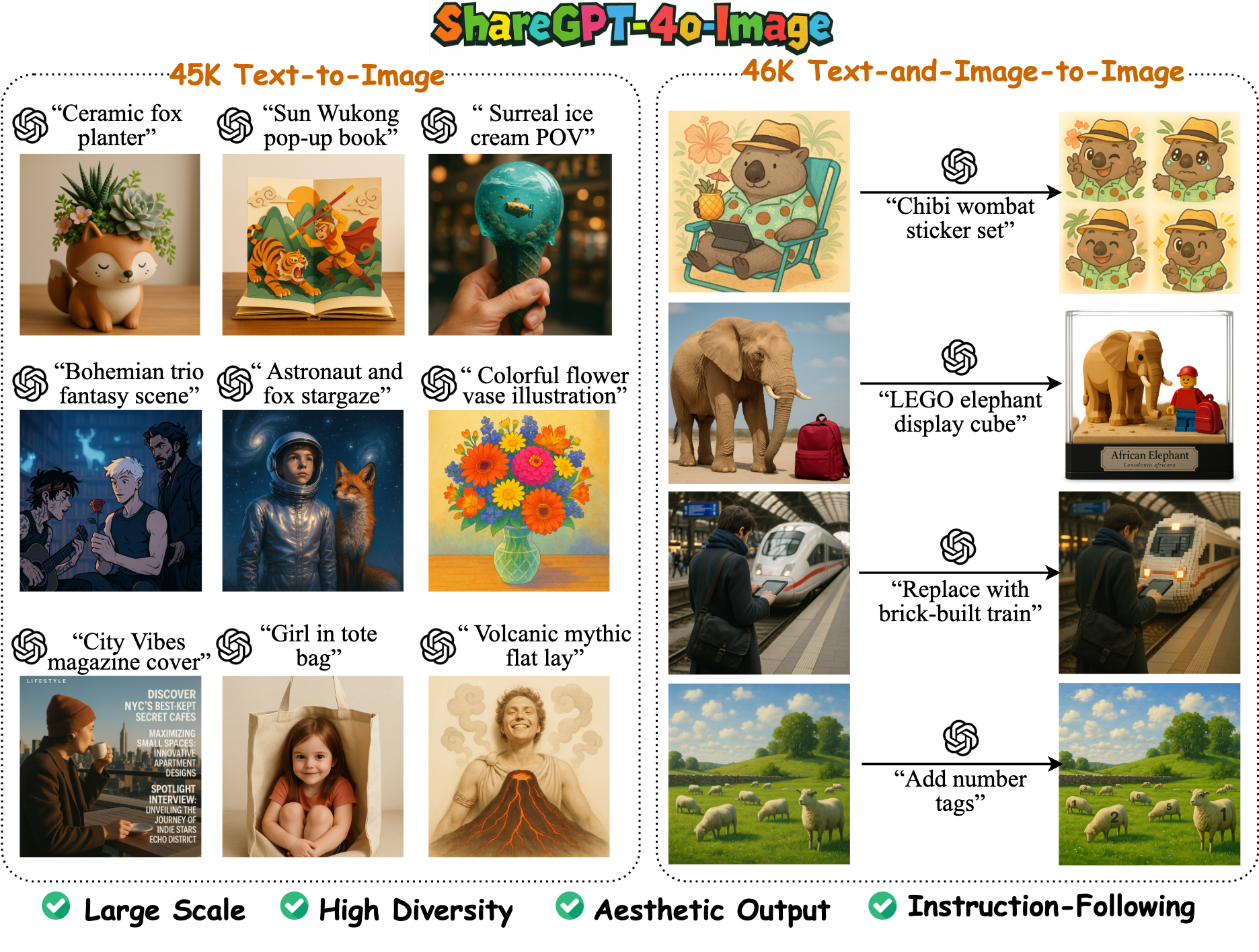}
  }
  \vspace{-2mm}
  \caption{\label{fig:head}Overview of the ShareGPT-4o-Image. The dataset comprises 91K synthetic samples from \textbf{GPT-4o-Image}, capturing its advanced capabilities for both \textit{text-to-image} and \textit{text-and-image-to-image} generation tasks. Displayed prompts are simplified.}
\end{figure*}

\begin{figure*}[ht!]
  \centering
  \resizebox{0.9\textwidth}{!}{
  \includegraphics[width=\textwidth]{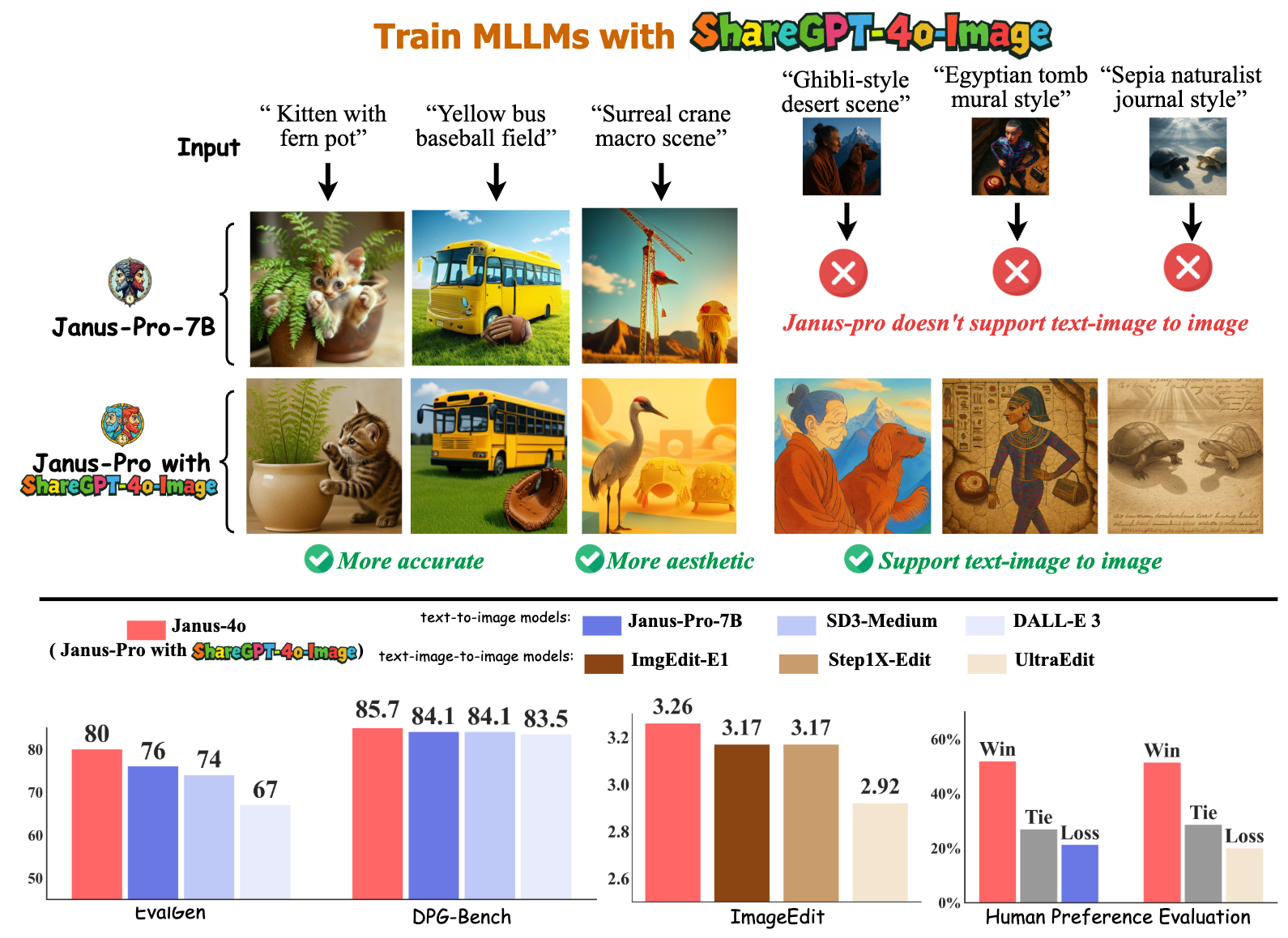}
  }
  \caption{\label{fig:head2}Image Generation Gains from ShareGPT-4o-Image. Fine-tuning Janus-Pro with ShareGPT-4o-Image yields \textbf{Janus-4o}, which shows notable improvements in image generation. Janus-4o also supports \textit{text-and-image-to-image} generation, outperforming other baselines with just 91K training samples.}
\end{figure*}

\section{Introduction}
Recent advancements in large-scale generative models~\citep{song2019generative,rombach2022high,ho2020denoising,ramesh2021zero,esser2021taming,rombach2022high,ho2022classifier} have significantly improved the quality and controllability of multimodal synthesis. Among these, image generation has emerged as a core capability in multimodal AI systems, enabling powerful tools for creative design, visual reasoning, and interactive agents. As large language models (LLMs) continue to evolve into natively multimodal architectures~\citep{achiam2023gpt,liu2023visual,qwen2.5vl,team2023gemini,chen2024internvl}, the boundary between language understanding and visual generation is rapidly dissolving, resulting in models that can follow complex instructions and produce highly contextualized, photorealistic images.

Recently, GPT-4o-Image~\citep{gpt4oimage}, the image generation variant of GPT-4o~\citep{gpt4o}, has demonstrated unprecedented capabilities in both \textit{text-to-image} and \textit{text-and-image-to-image} generation \citep{4oimagepowerful1}. It leverages a unified architecture that jointly learns from both modalities, resulting in superior instruction alignment, semantic coherence, and photorealism in image generation~\citep{gpt4oimage}. Notably, such proprietary models are characterized by highly opaque implementation processes, with even their architectures remaining undisclosed. In contrast, existing open-source image generation models, such as Janus-Pro~\citep{chen2025janus}, still fall significantly behind \citep{4oimagepowerful2}.

To make such capabilities more accessible to the community, we introduce \textbf{ShareGPT-4o-Image}, a synthesized dataset designed to transfer GPT-4o-level image generation skills to open multimodal models. The dataset includes 45K \textit{text-to-image} and 46K \textit{text-and-image-to-image} prompts, carefully curated for diversity and quality. Using GPT-4o's image generation, 
we synthesized the corresponding 91K image outputs generated by GPT-4o. The resulting dataset covers a wide range of styles and grounded visual reasoning, and reflects GPT-4o's strengths in instruction-following and visual aesthetics.

We fine-tune Janus-Pro~\citep{chen2025janus} on ShareGPT-4o-Image to develop \textbf{Janus-4o}, a new multimodal large language model (MLLM) that supports both \textit{text-to-image} and \textit{text-and-image-to-image} generation. In \textit{text-to-image} generation, Janus-4o improves on Janus-Pro by 4 and 1.6 points on EvalGen and DPG-Bench benchmarks, respectively.  Unlike its predecessor, which only handled \textit{text-to-image} tasks, Janus-4o introduces \textit{text-and-image-to-image} capabilities through innovative enhancements. On the ImgEdit-Bench, it excels among open-source image editing models, despite being trained on only 91K samples. Human evaluations also show a stronger preference for Janus-4o's outputs, highlighting the effectiveness of ShareGPT-4o-Image dataset. We are releasing both ShareGPT-4o-Image and Janus-4o to the research community to foster progress in aligning open-source multimodal models with cutting-edge generative capabilities.

Our contributions are summarized as follows:

\begin{itemize}
  \item We introduce ShareGPT-4o-Image, the first dataset comprising high-quality \textit{text-to-image} and \textit{text-and-image-to-image} pairs, distilled from GPT-4o image generation.
  \item Based on ShareGPT-4o-Image,  we develop Janus-4o, a multimodal large language model capable of both \textit{text-to-image} and \textit{text-and-image-to-image} generation.
  \item Experiments show that Janus-4o significantly improves upon Janus-Pro in image generation and enables \textit{text-and-image-to-image} generation. Remarkably, impressive performance in \textit{text-and-image-to-image} generation is achieved from scratch using only 91K synthetic samples and 6 hours of training on a single 8×A800 GPU machine.
\end{itemize}


\section{ShareGPT-4o-Image}

\begin{figure}[h!]
    \centering
    \includegraphics[width=0.95\linewidth]{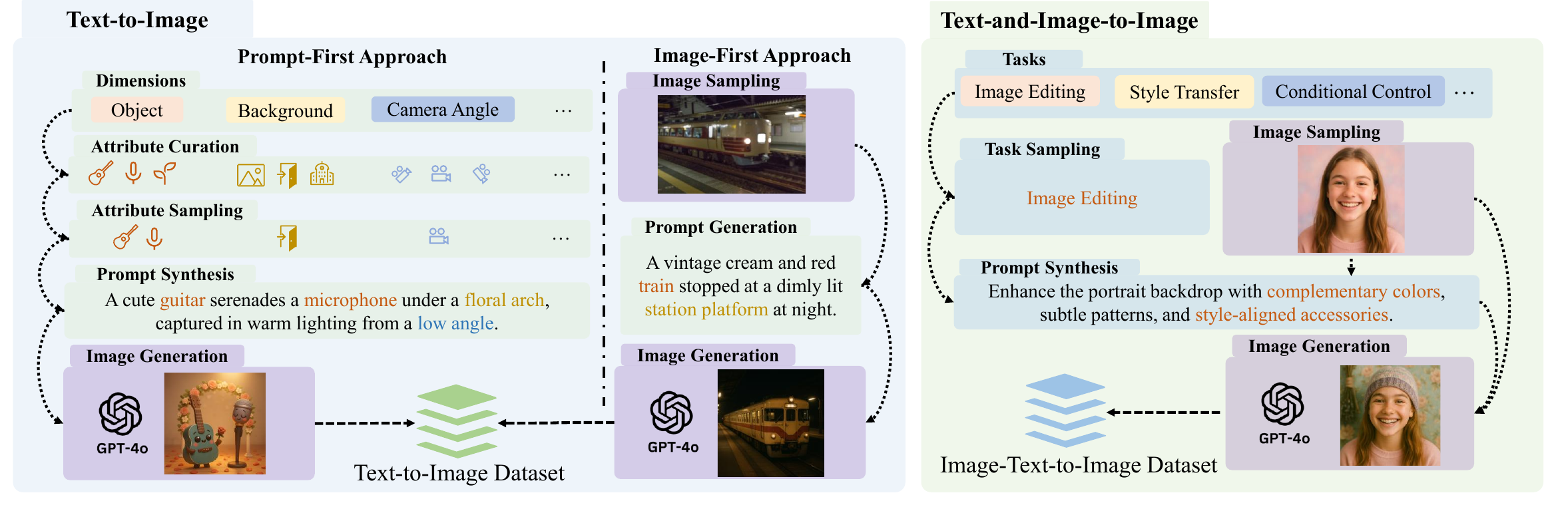}
    \caption{The flow diagram for the dataset construction process. The prompt in the diagram is the simplified version.}
    \label{fig:flow_diagram}
\end{figure}

To distill the advanced generative capabilities of GPT-4o-Image~\citep{gpt4oimage}, we construct \textbf{ShareGPT-4o-Image}, a large-scale dataset comprising 45k \textit{text-to-image} pairs and 46k instruction-guided image editing triplets. The entire data generation process, illustrated in Figure~\ref{fig:flow_diagram}, leverages Gemini-Pro-2.5 \citep{team2023gemini} as our primary Large Language Model (LLM) for all text synthesis tasks.

\subsection{\textit{Text-to-Image} Data}
\label{ssec:t2i_data}

We generate \textit{text-to-image} pairs via two complementary pipelines: a \textbf{prompt-first} approach for controlled attribute coverage and an \textbf{image-first} approach to ground prompts in realistic visual content.

\paragraph{Prompt-First Pipeline.}
This pipeline synthesizes structured prompts and then generates corresponding images. First, we define a six-dimensional attribute space (Objects, Background, Style, etc.) populated with a rich vocabulary, including 1,000 object categories from ImageNet (Appendix~\ref{app:category}). For each sample, we sample attributes from these dimensions and use our LLM to compose them into a coherent, natural-language prompt. Finally, each prompt is passed to GPT-4o-Image to generate a paired image. This systematic process yields diverse prompts with controlled complexity.

\paragraph{Image-First Pipeline.}
To complement the synthetic prompts, we source real-world, diverse, high-quality images from the ALLaVA dataset \citep{chen2024allava}. For each image, we prompt our LLM to generate a detailed, descriptive text prompt that accurately reflects the visual content (see Appendix~\ref{app:image_first_prompt} for the meta-prompt). The resulting (generated prompt, original image) pairs ensure our dataset's text distribution also captures the language needed to describe naturally occurring scenes.

\subsection{Instruction-Guided Image Editing Data}
\label{ssec:i2t2i_data}

To capture instruction-following editing capabilities, we generate data triplets of (source image, instruction, edited image). The generation process is as follows:

First, we define a taxonomy of 14 image editing tasks, grouped into 5 high-level categories such as object manipulation and style transfer (Appendix~\ref{app:image_text_task_categories}). For each data point, we begin with a source image, drawn either from our newly generated \textit{text-to-image} collection or from a curated set of real-world photos. We then sample an editing task from our taxonomy.

Based on the source image content and the selected task, our LLM synthesizes a specific, natural-language editing instruction (meta-prompts in Appendix~\ref{app:image_text_instruction_meta_prompts}). Finally, GPT-4o-Image executes this instruction on the source image to produce the edited output. This structured pipeline yields a diverse dataset of 46k triplets covering a wide range of common image editing scenarios.


\section{Janus-4o: Fine-Tuning with ShareGPT-4o-Image}
\begin{figure*}[h!]
  \centering
  \resizebox{1.0\textwidth}{!}{
  \includegraphics[width=\textwidth]{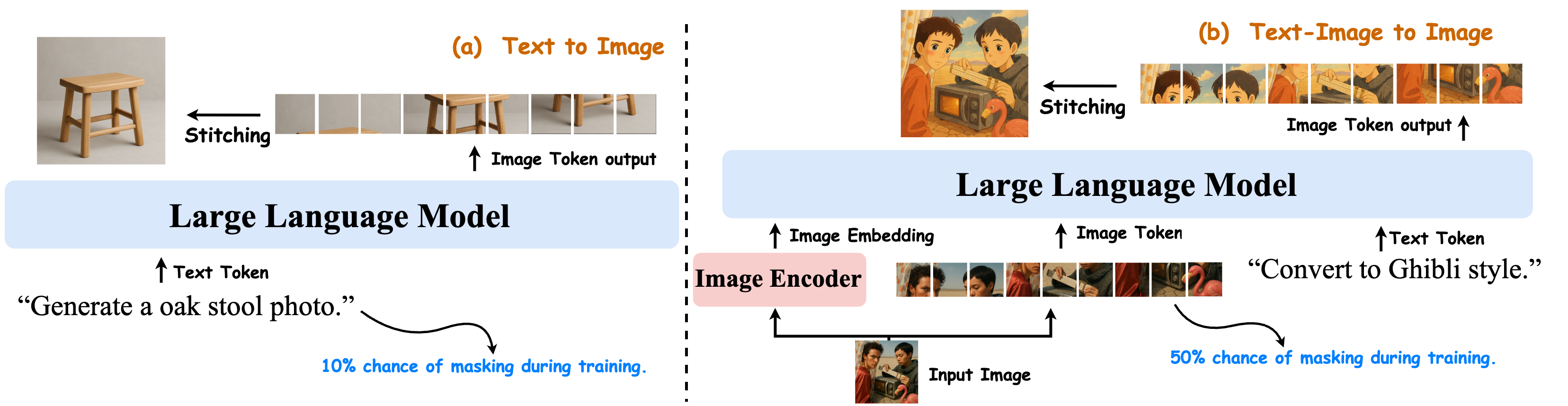}
  }
  \caption{\label{fig:method}Overview of Janus-4o model. Built upon Janus-Pro, it is constructed via fine-tuning on ShareGPT-4o-Image. It incorporates enhancements to support \textit{text-and-image-to-image} generation. Both \textit{text-to-image} and \textit{text-and-image-to-image} tasks are jointly trained.}
\end{figure*}

To assess the effectiveness of ShareGPT-4o-Image, we fine-tuned the ShareGPT-4o-Image dataset on Janus-Pro to develop \textbf{Janus-4o}.  The training methodology of Janus-4o is illustrated in Figure \ref{fig:method}.

\subsection{\texorpdfstring{\textit{Text-to-Image} Fine-Tuning}{Text-to-Image Fine-Tuning}}

\paragraph{Training} For \textit{text-to-image} training, we fine-tune Janus-Pro under the original \textit{text-to-image} generation setup. A text prompt is tokenized into $S = (s_0, s_1, \ldots, s_M)$, while the target image $I$ is flattened into pixel patches and mapped via a codebook into image tokens $X = (x_0, x_1, \ldots, x_N)$. Both $S$ and $X$ are embedded through their respective text and image embedding layers and then fed into the LLM. The objective is to autoregressively predict the image tokens, with the loss:
$$
\mathcal{L} = -\sum_{i=1} \log P_\theta\left(x_i \mid x_{<i}, S\right)
$$
During fine-tuning, 10\% of $S$ is randomly masked to padding tokens to encourage pixel-level dependency modeling, in line with GPT-4o-style modeling.

\paragraph{Inference} Inference follows the same next-token prediction process as Janus-Pro. For each token, the logit is computed as:

$$
l_g = l_u + s(l_c - l_u)
$$

where $l_c$ is the conditional logit (with prompt), $l_u$ the unconditional logit (prompt masked), and $s$ a scaling factor. We use $s = 5$ and a temperature of $1.0$.

\subsection{\textit{Text-and-Image-to-Image} Fine-Tuning: \textcolor{blue}{Unlocking New Capabilities}}
\paragraph{Training} Janus-Pro does not natively support \textit{text-and-image-to-image} generation. This task requires a semantic understanding of the input image to enable pixel-level modifications, making it necessary to incorporate both the image's semantic embedding and its tokenized representation.  Given an input image $\hat{I}$, an image encoder $\mathcal{E}$ produces a semantic embedding $\mathcal{E}(\hat{I})$, and a codebook yields image tokens $\hat{X} = (\hat{x}_0,\hat{x}_1,...,\hat{x}_N)$. These are concatenated with the prompt tokens $S$ to form the model input. The training loss for generating the target image $I$ is:

$$
\mathcal{L} = -\sum_{i=1} \log P_\theta\left(x_i \mid x_{<i},\mathcal{E}(\hat{I}), \hat{X}, S\right)
$$

To avoid overfitting to the input image, 50\% of $\hat{X}$ is randomly masked during fine-tuning.

 \paragraph{Inference}At inference, the input image is encoded into $\mathcal{E}(\hat{I})$ and $\hat{X}$, which are embedded into the input for autoregressive token generation. For each token, the logit is computed as:

$$
l_c' = \frac{l_c + s' \cdot l_o}{1 + s'}
$$
$$
l_g = l_u + s \cdot (l_c' - l_u)
$$

Here, $l_c$ is the conditional logit with full input, $l_o$ is the conditional logit with $\hat{X}$ masked, $l_u$ is the unconditional logit with all inputs masked, and $s = 5$, $s^\prime = 5$. The parameter $s'$ controls deviation from the input image: lower values preserve more of the original, while higher values allow more creative changes. The sampling temperature is set to $1.0$.

\subsection{Joint Fine-Tuning} 
\paragraph{Training Details} We fine-tune Janus-Pro-7B using the ShareGPT-4o-Image dataset, which includes 45K \textit{text-to-image} and 46K \textit{text-and-image-to-image} samples. \textbf{The two tasks are jointly trained via random sampling over 3 epochs.} All parameters of Janus-Pro-7B are fully fine-tuned, using a learning rate of $5 \times 10^{-6}$ and a batch size of 128. The resulting model, \textbf{Janus-4o}, supports both \textit{text-to-image} and \textit{text-and-image-to-image} generation tasks. \textbf{Training was completed in 6 hours on a single 8×A800 GPU machine.}


\section{Experiments}
\subsection{Experimental Setup}

\paragraph{Benchmarks}
To assess the \textit{text-to-image} generation capabilities of Janus-4o, we adopted the Janus-Pro evaluation protocol, leveraging GenEval~\citep{ghosh2024geneval} and DPG-Bench~\citep{hu2024ella} for a comprehensive analysis of compositionality and semantic alignment. In addition, we evaluated Janus-4o's ability to generate images conditioned on both visual and textual inputs using ImgEdit~\citep{ye2025imgedit}, focusing on single-turn edits.

\paragraph{Human Evaluation} To evaluate alignment with human preferences, we conducted a human study comparing Janus-4o against Janus-Pro-7B~\citep{chen2025janus} and UltraEdit~\citep{ultraedit}. The comparison involved 52 \textit{Text-to-Image} and 35 \textit{Text-and-Image-to-Image} examples sourced from real Twitter posts. Model outputs were randomly ordered, and a single evaluator selected the preferred output based on instruction fidelity and visual clarity, or marked a tie. Results were reported as preference ratios.

\subsection{Results and Analysis}

\begin{table}[ht]
\centering
\setlength{\tabcolsep}{4pt}
\renewcommand{\arraystretch}{1.2}
\resizebox{0.9\textwidth}{!}{
\begin{tabular}{c|cccccc|c}
\toprule
\textbf{Method}  & \textbf{Single Obj.} & \textbf{Two Obj.} & \textbf{Counting} & \textbf{Colors} & \textbf{Position} & \textbf{Color Attri.} & \textbf{Overall$\uparrow$} \\
\midrule
\multicolumn{8}{c}{\textbf{\textit{Text-to-Image} Generation Models}} \\ \hline
LlamaGen~\citep{llamagen}  & 0.71 & 0.34 & 0.21 & 0.58 & 0.07 & 0.04 & 0.32 \\
LDM~\citep{rombach2022high} & 0.92 & 0.29 & 0.23 & 0.70 & 0.02 & 0.05 & 0.37 \\
SDv$1.5$~\citep{rombach2022high} &  0.97 & 0.38 & 0.35 & 0.76 & 0.04 & 0.06 & 0.43 \\
PixArt-$\alpha$~\citep{chen2023pixart} &  0.98 & 0.50 & 0.44 & 0.80 & 0.08 & 0.07 & 0.48 \\
SDv$2.1$~\citep{rombach2022high} &  0.98 & 0.51 & 0.44 & 0.85 & 0.07 & 0.17 & 0.50 \\
DALL-E $2$~\citep{ramesh2022hierarchical}  & 0.94 & 0.66 & 0.49 & 0.77 & 0.10 & 0.19 & 0.52 \\
Emu$3$-Gen~\citep{wang2024emu3}  & 0.98 & 0.71 & 0.34 & 0.81 & 0.17 & 0.21 & 0.54 \\
SDXL~\citep{podell2023sdxl} &  0.98 & 0.74 & 0.39 & 0.85 & 0.15 & 0.23 & 0.55 \\
DALL-E $3$~\citep{dalle3}  & 0.96 & 0.87 & 0.47 & 0.83 & 0.43 & 0.45 & 0.67 \\
SD3-Medium~\citep{esser2024scalingrectifiedflowtransformers} & 0.99 & \textbf{0.94} & \textbf{0.72} & 0.89 & 0.33 & 0.60 & 0.74 \\
Janus-Pro-7B$^\dagger$\citep{chen2025janus}& \textbf{1.00} & 0.85 & 0.53 & \textbf{0.90} & 0.69 & 0.58 &  0.76 \\
\hline
\multicolumn{8}{c}{\textbf{Our Multimodal LLM}} \\ \hline
 \makecell[c]{ \textbf{Janus-4o}  \\  (Janus-Pro + \raisebox{-0.2em}{\includegraphics[width=1in]{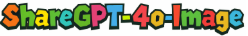}})}  & \textbf{1.00} & 0.92 & 0.58 & 0.88 & \textbf{0.70} & \textbf{0.70} & \textbf{0.80} \\
\bottomrule
\end{tabular}
}
\caption{\label{tab:geneval}\textbf{Evaluation of \textit{text-to-image} generation ability on GenEval benchmark}. $^\dagger$ indicates results rigorously reproduced by us, while others are taken from the original papers.}
\end{table}

\begin{table}[ht]
\centering
\renewcommand{\arraystretch}{1.2}
\resizebox{0.8\textwidth}{!}{
\begin{tabular}{c|ccccc|c}
\toprule
\textbf{Method} & \textbf{Global} & \textbf{Entity} & \textbf{Attribute} & \textbf{Relation} & \textbf{Other} & \textbf{Overall$\uparrow$} \\
\midrule
\multicolumn{7}{c}{\textbf{\textit{Text-to-Image} Generation Models}} \\ \hline
SDv1.5 \citep{rombach2022high} & 74.63 & 74.23 & 75.39 & 73.49 & 67.81 & 63.18 \\
PixArt-$\alpha$ \citep{chen2023pixart} & 74.97 & 79.32 & 78.60 & 82.57 & 76.96 & 71.11 \\
Lumina-Next \citep{2024lumina} & 82.82 & 88.65 & 86.44 & 80.53 & 81.82 & 74.63 \\
SDXL \citep{podell2023sdxl} & 83.27 & 82.43 & 80.91 & 86.76 & 80.41 & 74.65 \\
Playground v2.5 \citep{2024PG2.5} & 83.06 & 82.59 & 81.20 & 84.08 & 83.50 & 75.47 \\
Hunyuan-DiT \citep{2024hunyuandit} & 84.59 & 80.59 & 88.01 & 74.36 & 86.41 & 78.87 \\
PixArt-$\Sigma$ \citep{2024pixartsigma} & 86.89 & 82.89 & 88.94 & 86.59 & 87.68 & 80.54\\
Emu3-Gen \citep{wang2024emu3} & 85.21 & 86.68 & 86.84 & 90.22 & 83.15 & 80.60 \\
DALL-E 3 \citep{dalle3} & 90.97 & 89.61 & 88.39 & 90.58 & \textbf{89.83} & 83.50 \\
SD3-Medium \citep{esser2024scalingrectifiedflowtransformers} & 87.90 & \textbf{91.01} & 88.83 & 80.70 & 88.68 & 84.08 \\
Janus-Pro-7B$^\dagger$\citep{chen2025janus} & 80.70 & 89.75 & \textbf{90.04} & 89.85 & 88.55 & 84.12 \\
\hline
\multicolumn{7}{c}{\textbf{Our Multimodal LLM}} \\ \hline
 \makecell[c]{ \textbf{Janus-4o}  \\  (Janus-Pro + \raisebox{-0.2em}{\includegraphics[width=1in]{image/logo.png}})} & \textbf{92.59} & 90.61 & 89.51 & \textbf{91.77} & 89.01 & \textbf{85.71} \\
\bottomrule
\end{tabular}
}
\caption{\label{tab:exp-dpg}\textbf{Evaluation of \textit{text-to-image} generation ability on DPG-Bench benchmark}. $^\dagger$ indicates results rigorously reproduced by us, while others are reported in the respective papers.}
\end{table}

\paragraph{\textit{Text-to-Image} Performance}
We evaluate the performance of \textit{text-to-image} generation on GenEval and DPG-Bench. As shown in Table \ref{tab:geneval}, Janus-4o achieves a 4-point improvement over Janus-Pro, highlighting the benefit of the ShareGPT-4o-Image. Its 80\% overall accuracy also surpasses other baseline models. In Table \ref{tab:exp-dpg}, Janus-4o achieves a score of 85.71 on DPG-Bench, representing a 1.6-point improvement over Janus-Pro. These demonstrate that ShareGPT-4o-Image not only enhances image generation quality but also improves the model's ability to follow dense instructions for \textit{text-to-image} generation.

\begin{table}[thbp]
\centering
\normalsize
\resizebox{1\textwidth}{!}{
\begin{tabular}{c|c|cccccccc|c}
\toprule
\textbf{Model} & \textbf{\#Data$\downarrow$} & \textbf{Add.} & \textbf{Rmv.} & \textbf{Repl.} & \textbf{Mot.} & \textbf{Style} & \textbf{Bkg.} & \textbf{Obj.} & \textbf{Hyb.} & \textbf{Avg.$\uparrow$} \\
\midrule
\multicolumn{11}{c}{\textbf{\textit{Text-and-Image-to-Image} Generation Models}} \\ \midrule
\makecell[c]{ Instruct-Pix2Pix  \\\citep{instructp2p} } & 500K & 2.29 & 1.49 & 1.93 & 1.51 & 3.54 & 1.67 & 1.33 & 1.48 & 1.91 \\
AnySD~\citep{anyedit} & 2500K & 3.12 & 2.34 & 2.71 & 3.31 & 3.27 & 2.37 & 1.82 & 2.07 & 2.62 \\
UltraEdit~\citep{ultraedit} & 4100K & 3.63 & 1.71 & 3.13 & 3.57 & 3.69 & 3.31 & 2.02 & 2.33 & 2.92 \\
Step1X-Edit~\citep{liu2025step1x} & 1000K & \textbf{3.90} & \textbf{2.61} & \textbf{3.45} & 3.43 & 4.44 & 3.19 & 1.87 & 2.52 & 3.17 \\
ImgEdit-E1~\citep{ye2025imgedit} & 1200K & 3.82 & 2.40 & 2.80 & 3.21 & 4.38 & \textbf{3.38} & \textbf{2.55} & \textbf{2.87} & 3.17 \\
\midrule
\multicolumn{11}{c}{\textbf{Our Multimodal LLM}} \\ \hline
 \makecell[c]{ \textbf{Janus-4o}  \\  (Janus-Pro + \raisebox{-0.2em}{\includegraphics[width=1in]{image/logo.png}})} & \textbf{91K} & 3.60 & 2.28 & 3.27 & \textbf{4.13} & \textbf{4.47} & 3.32 & 2.28 & 2.74 & \textbf{3.26} \\
\bottomrule
\end{tabular}
}
\caption{\label{tab:imgedit}\textbf{Evaluation of \textit{text-and-image-to-image} generation ability on ImgEdit-Bench.} \textbf{\#Data} denotes training data size (K = thousand). Column abbreviations: Add. = Addition, Rmv. = Removement, Repl. = Replacement, Mot. = Motion Change, Style = Style Transfer, Bkg. = Background Change, Obj. = Object Extraction, Hyb. = Hybrid Edit, Avg. = Average across all edits.}
\end{table}

\paragraph{\textit{Text-and-Image-to-Image} Performance}
We evaluate the performance of \textit{text-and-image-to-image} generation on ImgEdit-Bench. As shown in Table \ref{tab:imgedit}, Janus-4o achieved a score of 3.26, surpassing other baselines like Step1X-Edit and ImgEdit-E1, with particularly strong results in the \textit{Motion Change} and \textit{Style Transfer} categories. Remarkably, Janus-4o attained competitive performance with only 91K training samples, despite the original model lacking support for \textit{text-and-image-to-image} generation. This underscores the critical role of high-quality data and highlights the effectiveness of the ShareGPT-4o-Image dataset.

\begin{figure*}[h!]
  \centering
  \resizebox{0.65\textwidth}{!}{
  \includegraphics[width=\textwidth]{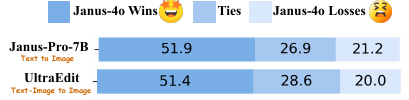}
  }
  \caption{Human evaluation of Janus-4o's performance on \textit{text-to-image} and \textit{text-and-image-to-image} generation tasks. \textit{Wins} indicate that the evaluator found Janus-4o superior in both instruction-following and image quality. \textit{Ties} mean both models performed equally well, while \textit{Losses} indicate that Janus-4o performed worse in those cases. Results were reported as preference ratios.}
  \label{fig: humaneval}
\end{figure*}

\paragraph{Human Evaluation} The results in Figure \ref{fig: humaneval} indicate that content generated by Janus-4o aligns more closely with human preferences than that of previous models. In the \textit{Text-to-Image} task, Janus-4o produces images with noticeably higher visual quality compared to Janus-Pro. Furthermore, in the \textit{Text-and-Image-to-Image} task, it exhibits a significantly improved ability to follow instructions, leading to more accurate image modifications based on the given prompts.

\section{conclusion}
In this work, we introduce ShareGPT-4o-Image, the first large-scale dataset that captures GPT-4o's advanced image generation capabilities in both \textit{text-to-image} and \textit{text-and-image-to-image} generation. Using this dataset, we develop Janus-4o, a MLLM capable of generating high-quality images from text alone or from combined image and text inputs. Janus-4o achieves substantial improvements in \textit{text-to-image} generation and delivers competitive results on \textit{text-and-image-to-image} tasks, underscoring the high quality and utility of ShareGPT-4o-Image. Benefiting from the efficiency of MLLM-based autoregressive image generation, Janus-4o can be trained in just 6 hours on an 8×A800 GPU machine, achieving notable performance gains with minimal compute.


\bibliography{colm2025_conference}

\begin{thebibliography}{65}
\providecommand{\natexlab}[1]{#1}
\providecommand{\url}[1]{\texttt{#1}}
\expandafter\ifx\csname urlstyle\endcsname\relax
  \providecommand{\doi}[1]{doi: #1}\else
  \providecommand{\doi}{doi: \begingroup \urlstyle{rm}\Url}\fi

\bibitem[Achiam et~al.(2023)Achiam, Adler, Agarwal, Ahmad, Akkaya, Aleman, Almeida, Altenschmidt, Altman, Anadkat, et~al.]{achiam2023gpt}
Josh Achiam, Steven Adler, Sandhini Agarwal, Lama Ahmad, Ilge Akkaya, Florencia~Leoni Aleman, Diogo Almeida, Janko Altenschmidt, Sam Altman, Shyamal Anadkat, et~al.
\newblock Gpt-4 technical report.
\newblock \emph{arXiv preprint arXiv:2303.08774}, 2023.

\bibitem[Bai et~al.(2025)Bai, Chen, Liu, Wang, Ge, Song, Dang, Wang, Wang, Tang, et~al.]{qwen2.5vl}
Shuai Bai, Keqin Chen, Xuejing Liu, Jialin Wang, Wenbin Ge, Sibo Song, Kai Dang, Peng Wang, Shijie Wang, Jun Tang, et~al.
\newblock Qwen2. 5-vl technical report.
\newblock \emph{arXiv preprint arXiv:2502.13923}, 2025.

\bibitem[Betker et~al.(2023)Betker, Goh, Jing, Brooks, Wang, Li, Ouyang, Zhuang, Lee, Guo, et~al.]{dalle3}
James Betker, Gabriel Goh, Li~Jing, Tim Brooks, Jianfeng Wang, Linjie Li, Long Ouyang, Juntang Zhuang, Joyce Lee, Yufei Guo, et~al.
\newblock Improving image generation with better captions.
\newblock \emph{Computer Science. https://cdn. openai. com/papers/dall-e-3. pdf}, 2\penalty0 (3):\penalty0 8, 2023.

\bibitem[Brooks et~al.(2022)Brooks, Holynski, and Efros]{instructp2p}
Tim Brooks, Aleksander Holynski, and Alexei~A. Efros.
\newblock Instructpix2pix: Learning to follow image editing instructions.
\newblock \emph{2023 IEEE/CVF Conference on Computer Vision and Pattern Recognition (CVPR)}, pp.\  18392--18402, 2022.
\newblock URL \url{https://api.semanticscholar.org/CorpusID:253581213}.

\bibitem[Cai et~al.(2025)Cai, Chen, Chen, Li, Long, Pan, Qiu, Zhang, Gao, Xu, et~al.]{cai2025hidream}
Qi~Cai, Jingwen Chen, Yang Chen, Yehao Li, Fuchen Long, Yingwei Pan, Zhaofan Qiu, Yiheng Zhang, Fengbin Gao, Peihan Xu, et~al.
\newblock Hidream-i1: A high-efficient image generative foundation model with sparse diffusion transformer.
\newblock \emph{arXiv preprint arXiv:2505.22705}, 2025.

\bibitem[Changpinyo et~al.(2021)Changpinyo, Sharma, Ding, and Soricut]{cc12m}
Soravit Changpinyo, Piyush Sharma, Nan Ding, and Radu Soricut.
\newblock Conceptual 12m: Pushing web-scale image-text pre-training to recognize long-tail visual concepts.
\newblock In \emph{Proceedings of the IEEE/CVF conference on computer vision and pattern recognition}, pp.\  3558--3568, 2021.

\bibitem[Chen et~al.(2024{\natexlab{a}})Chen, Chen, Zhang, Chen, Wu, Zhang, Chen, Li, Wan, and Wang]{chen2024allava}
Guiming~Hardy Chen, Shunian Chen, Ruifei Zhang, Junying Chen, Xiangbo Wu, Zhiyi Zhang, Zhihong Chen, Jianquan Li, Xiang Wan, and Benyou Wang.
\newblock Allava: Harnessing gpt4v-synthesized data for lite vision-language models.
\newblock \emph{arXiv preprint arXiv:2402.11684}, 2024{\natexlab{a}}.

\bibitem[Chen et~al.(2024{\natexlab{b}})Chen, Huang, Lv, Cui, Chen, and Wei]{textdiffuser}
Jingye Chen, Yupan Huang, Tengchao Lv, Lei Cui, Qifeng Chen, and Furu Wei.
\newblock Textdiffuser: Diffusion models as text painters.
\newblock \emph{Advances in Neural Information Processing Systems}, 36, 2024{\natexlab{b}}.

\bibitem[Chen et~al.(2025{\natexlab{a}})Chen, Xu, Pan, Hu, Qin, Goldstein, Huang, Zhou, Xie, Savarese, et~al.]{chen2025blip3o}
Jiuhai Chen, Zhiyang Xu, Xichen Pan, Yushi Hu, Can Qin, Tom Goldstein, Lifu Huang, Tianyi Zhou, Saining Xie, Silvio Savarese, et~al.
\newblock Blip3-o: A family of fully open unified multimodal models-architecture, training and dataset.
\newblock \emph{arXiv preprint arXiv:2505.09568}, 2025{\natexlab{a}}.

\bibitem[Chen et~al.(2023)Chen, Yu, Ge, Yao, Xie, Wu, Wang, Kwok, Luo, Lu, et~al.]{chen2023pixart}
Junsong Chen, Jincheng Yu, Chongjian Ge, Lewei Yao, Enze Xie, Yue Wu, Zhongdao Wang, James Kwok, Ping Luo, Huchuan Lu, et~al.
\newblock Pixart-$alpha$: Fast training of diffusion transformer for photorealistic text-to-image synthesis.
\newblock \emph{arXiv preprint arXiv:2310.00426}, 2023.

\bibitem[Chen et~al.(2024{\natexlab{c}})Chen, Ge, Xie, Wu, Yao, Ren, Wang, Luo, Lu, and Li]{2024pixartsigma}
Junsong Chen, Chongjian Ge, Enze Xie, Yue Wu, Lewei Yao, Xiaozhe Ren, Zhongdao Wang, Ping Luo, Huchuan Lu, and Zhenguo Li.
\newblock {PixArt-Sigma}: Weak-to-strong training of diffusion transformer for {4K} text-to-image generation.
\newblock \emph{arXiv preprint arXiv:2403.04692}, 2024{\natexlab{c}}.

\bibitem[Chen et~al.(2024{\natexlab{d}})Chen, Gui, Ouyang, Gao, Chen, Chen, Wang, Cai, Ji, Wan, et~al.]{chen2024towards}
Junying Chen, Chi Gui, Ruyi Ouyang, Anningzhe Gao, Shunian Chen, Guiming Chen, Xidong Wang, Zhenyang Cai, Ke~Ji, Xiang Wan, et~al.
\newblock Towards injecting medical visual knowledge into multimodal llms at scale.
\newblock In \emph{Proceedings of the 2024 Conference on Empirical Methods in Natural Language Processing}, pp.\  7346--7370, 2024{\natexlab{d}}.

\bibitem[Chen et~al.(2025{\natexlab{b}})Chen, Bai, Zhao, Ye, Shi, Zhou, Chai, Lin, Wu, Tang, et~al.]{4oimagepowerful1}
Sixiang Chen, Jinbin Bai, Zhuoran Zhao, Tian Ye, Qingyu Shi, Donghao Zhou, Wenhao Chai, Xin Lin, Jianzong Wu, Chao Tang, et~al.
\newblock An empirical study of gpt-4o image generation capabilities.
\newblock \emph{arXiv preprint arXiv:2504.05979}, 2025{\natexlab{b}}.

\bibitem[Chen et~al.(2025{\natexlab{c}})Chen, Wu, Liu, Pan, Liu, Xie, Yu, and Ruan]{chen2025janus}
Xiaokang Chen, Zhiyu Wu, Xingchao Liu, Zizheng Pan, Wen Liu, Zhenda Xie, Xingkai Yu, and Chong Ruan.
\newblock Janus-pro: Unified multimodal understanding and generation with data and model scaling.
\newblock \emph{arXiv preprint arXiv:2501.17811}, 2025{\natexlab{c}}.

\bibitem[Chen et~al.(2024{\natexlab{e}})Chen, Wu, Wang, Su, Chen, Xing, Zhong, Zhang, Zhu, Lu, et~al.]{chen2024internvl}
Zhe Chen, Jiannan Wu, Wenhai Wang, Weijie Su, Guo Chen, Sen Xing, Muyan Zhong, Qinglong Zhang, Xizhou Zhu, Lewei Lu, et~al.
\newblock Internvl: Scaling up vision foundation models and aligning for generic visual-linguistic tasks.
\newblock In \emph{Proceedings of the IEEE/CVF Conference on Computer Vision and Pattern Recognition}, pp.\  24185--24198, 2024{\natexlab{e}}.

\bibitem[Deng et~al.(2009)Deng, Dong, Socher, Li, Li, and Fei-Fei]{deng2009imagenet}
Jia Deng, Wei Dong, Richard Socher, Li-Jia Li, Kai Li, and Li~Fei-Fei.
\newblock Imagenet: A large-scale hierarchical image database.
\newblock In \emph{2009 IEEE conference on computer vision and pattern recognition}, pp.\  248--255. Ieee, 2009.

\bibitem[Esser et~al.(2021)Esser, Rombach, and Ommer]{esser2021taming}
Patrick Esser, Robin Rombach, and Bjorn Ommer.
\newblock Taming transformers for high-resolution image synthesis.
\newblock In \emph{Proceedings of the IEEE/CVF conference on computer vision and pattern recognition}, pp.\  12873--12883, 2021.

\bibitem[Esser et~al.(2024)Esser, Kulal, Blattmann, Entezari, Müller, Saini, Levi, Lorenz, Sauer, Boesel, Podell, Dockhorn, English, Lacey, Goodwin, Marek, and Rombach]{esser2024scalingrectifiedflowtransformers}
Patrick Esser, Sumith Kulal, Andreas Blattmann, Rahim Entezari, Jonas Müller, Harry Saini, Yam Levi, Dominik Lorenz, Axel Sauer, Frederic Boesel, Dustin Podell, Tim Dockhorn, Zion English, Kyle Lacey, Alex Goodwin, Yannik Marek, and Robin Rombach.
\newblock Scaling rectified flow transformers for high-resolution image synthesis, 2024.
\newblock URL \url{https://arxiv.org/abs/2403.03206}.

\bibitem[Fei et~al.(2024)Fei, Wu, Zhang, Chua, and Yan]{fei2024vitron}
Hao Fei, Shengqiong Wu, Hanwang Zhang, Tat-Seng Chua, and Shuicheng Yan.
\newblock Vitron: A unified pixel-level vision llm for understanding, generating, segmenting, editing.
\newblock In \emph{Advances in Neural Information Processing Systems}, 2024.

\bibitem[Ge et~al.(2024{\natexlab{a}})Ge, Zhao, Li, Ge, and Shan]{seeddataedit}
Yuying Ge, Sijie Zhao, Chen Li, Yixiao Ge, and Ying Shan.
\newblock Seed-data-edit technical report: A hybrid dataset for instructional image editing.
\newblock \emph{arXiv preprint arXiv:2405.04007}, 2024{\natexlab{a}}.

\bibitem[Ge et~al.(2024{\natexlab{b}})Ge, Zhao, Zhu, Ge, Yi, Song, Li, Ding, and Shan]{ge2024seed}
Yuying Ge, Sijie Zhao, Jinguo Zhu, Yixiao Ge, Kun Yi, Lin Song, Chen Li, Xiaohan Ding, and Ying Shan.
\newblock Seed-x: Multimodal models with unified multi-granularity comprehension and generation.
\newblock \emph{arXiv preprint arXiv:2404.14396}, 2024{\natexlab{b}}.

\bibitem[Ghosh et~al.(2024)Ghosh, Hajishirzi, and Schmidt]{ghosh2024geneval}
Dhruba Ghosh, Hannaneh Hajishirzi, and Ludwig Schmidt.
\newblock Geneval: An object-focused framework for evaluating text-to-image alignment.
\newblock \emph{Advances in Neural Information Processing Systems}, 36, 2024.

\bibitem[Henighan et~al.(2020)Henighan, Kaplan, Katz, Chen, Hesse, Jackson, Jun, Brown, Dhariwal, Gray, et~al.]{henighan2020scaling}
Tom Henighan, Jared Kaplan, Mor Katz, Mark Chen, Christopher Hesse, Jacob Jackson, Heewoo Jun, Tom~B Brown, Prafulla Dhariwal, Scott Gray, et~al.
\newblock Scaling laws for autoregressive generative modeling.
\newblock \emph{arXiv preprint arXiv:2010.14701}, 2020.

\bibitem[Ho \& Salimans(2022)Ho and Salimans]{ho2022classifier}
Jonathan Ho and Tim Salimans.
\newblock Classifier-free diffusion guidance.
\newblock \emph{arXiv preprint arXiv:2207.12598}, 2022.

\bibitem[Ho et~al.(2020)Ho, Jain, and Abbeel]{ho2020denoising}
Jonathan Ho, Ajay Jain, and Pieter Abbeel.
\newblock Denoising diffusion probabilistic models.
\newblock \emph{Advances in neural information processing systems}, 33:\penalty0 6840--6851, 2020.

\bibitem[Hu et~al.(2024)Hu, Wang, Fang, Fu, Cheng, and Yu]{hu2024ella}
Xiwei Hu, Rui Wang, Yixiao Fang, Bin Fu, Pei Cheng, and Gang Yu.
\newblock Ella: Equip diffusion models with llm for enhanced semantic alignment.
\newblock \emph{arXiv preprint arXiv:2403.05135}, 2024.

\bibitem[Hui et~al.(2024)Hui, Yang, Zhao, Shi, Wang, Wang, Zhou, and Xie]{hqedit}
Mude Hui, Siwei Yang, Bingchen Zhao, Yichun Shi, Heng Wang, Peng Wang, Yuyin Zhou, and Cihang Xie.
\newblock Hq-edit: A high-quality dataset for instruction-based image editing.
\newblock \emph{arXiv preprint arXiv:2404.09990}, 2024.

\bibitem[Li et~al.(2024{\natexlab{a}})Li, Kamko, Akhgari, Sabet, Xu, and Doshi]{2024PG2.5}
Daiqing Li, Aleks Kamko, Ehsan Akhgari, Ali Sabet, Linmiao Xu, and Suhail Doshi.
\newblock Playground v2.5: Three insights towards enhancing aesthetic quality in text-to-image generation.
\newblock \emph{arXiv preprint arXiv:2402.17245}, 2024{\natexlab{a}}.

\bibitem[Li et~al.(2024{\natexlab{b}})Li, Zhang, Lin, Xiong, Long, Deng, Zhang, Liu, Huang, Xiao, et~al.]{2024hunyuandit}
Zhimin Li, Jianwei Zhang, Qin Lin, Jiangfeng Xiong, Yanxin Long, Xinchi Deng, Yingfang Zhang, Xingchao Liu, Minbin Huang, Zedong Xiao, et~al.
\newblock {Hunyuan-DiT}: A powerful multi-resolution diffusion transformer with fine-grained chinese understanding.
\newblock \emph{arXiv preprint arXiv:2405.08748}, 2024{\natexlab{b}}.

\bibitem[Lin et~al.(2014)Lin, Maire, Belongie, Hays, Perona, Ramanan, Doll{\'a}r, and Zitnick]{mscoco}
Tsung-Yi Lin, Michael Maire, Serge Belongie, James Hays, Pietro Perona, Deva Ramanan, Piotr Doll{\'a}r, and C~Lawrence Zitnick.
\newblock Microsoft coco: Common objects in context.
\newblock In \emph{Computer Vision--ECCV 2014: 13th European Conference, Zurich, Switzerland, September 6-12, 2014, Proceedings, Part V 13}, pp.\  740--755. Springer, 2014.

\bibitem[Liu et~al.(2024)Liu, Yan, Zaharia, and Abbeel]{liu2024world}
Hao Liu, Wilson Yan, Matei Zaharia, and Pieter Abbeel.
\newblock World model on million-length video and language with ringattention.
\newblock \emph{arXiv preprint arXiv:2402.08268}, 2024.

\bibitem[Liu et~al.(2023)Liu, Li, Wu, and Lee]{liu2023visual}
Haotian Liu, Chunyuan Li, Qingyang Wu, and Yong~Jae Lee.
\newblock Visual instruction tuning.
\newblock \emph{Advances in neural information processing systems}, 36:\penalty0 34892--34916, 2023.

\bibitem[Liu et~al.(2025)Liu, Han, Xing, Yin, Wang, Cheng, Liao, Wang, Fu, Han, et~al.]{liu2025step1x}
Shiyu Liu, Yucheng Han, Peng Xing, Fukun Yin, Rui Wang, Wei Cheng, Jiaqi Liao, Yingming Wang, Honghao Fu, Chunrui Han, et~al.
\newblock Step1x-edit: A practical framework for general image editing.
\newblock \emph{arXiv preprint arXiv:2504.17761}, 2025.

\bibitem[Lu et~al.(2024)Lu, Clark, Lee, Zhang, Khosla, Marten, Hoiem, and Kembhavi]{unifiedio2}
Jiasen Lu, Christopher Clark, Sangho Lee, Zichen Zhang, Savya Khosla, Ryan Marten, Derek Hoiem, and Aniruddha Kembhavi.
\newblock Unified-io 2: Scaling autoregressive multimodal models with vision language audio and action.
\newblock In \emph{Proceedings of the IEEE/CVF Conference on Computer Vision and Pattern Recognition}, pp.\  26439--26455, 2024.

\bibitem[{Open AI}(2025)]{gpt4oimage}
{Open AI}.
\newblock Introducing 4o image generation.
\newblock \url{https://openai.com/index/introducing-4o-image-generation/}, 2025.

\bibitem[OpenAI(2024)]{gpt4o}
OpenAI.
\newblock Hello gpt-4o, 2024.
\newblock URL \url{https://openai.com/index/hello-gpt-4o}.
\newblock Accessed: 2024-09-09.

\bibitem[Podell et~al.(2023)Podell, English, Lacey, Blattmann, Dockhorn, M{\"u}ller, Penna, and Rombach]{podell2023sdxl}
Dustin Podell, Zion English, Kyle Lacey, Andreas Blattmann, Tim Dockhorn, Jonas M{\"u}ller, Joe Penna, and Robin Rombach.
\newblock Sdxl: Improving latent diffusion models for high-resolution image synthesis.
\newblock \emph{arXiv preprint arXiv:2307.01952}, 2023.

\bibitem[Ramesh et~al.(2021)Ramesh, Pavlov, Goh, Gray, Voss, Radford, Chen, and Sutskever]{ramesh2021zero}
Aditya Ramesh, Mikhail Pavlov, Gabriel Goh, Scott Gray, Chelsea Voss, Alec Radford, Mark Chen, and Ilya Sutskever.
\newblock Zero-shot text-to-image generation.
\newblock In \emph{International conference on machine learning}, pp.\  8821--8831. Pmlr, 2021.

\bibitem[Ramesh et~al.(2022)Ramesh, Dhariwal, Nichol, Chu, and Chen]{ramesh2022hierarchical}
Aditya Ramesh, Prafulla Dhariwal, Alex Nichol, Casey Chu, and Mark Chen.
\newblock Hierarchical text-conditional image generation with clip latents.
\newblock \emph{arXiv preprint arXiv:2204.06125}, 1\penalty0 (2):\penalty0 3, 2022.

\bibitem[Rombach et~al.(2022)Rombach, Blattmann, Lorenz, Esser, and Ommer]{rombach2022high}
Robin Rombach, Andreas Blattmann, Dominik Lorenz, Patrick Esser, and Bj{\"o}rn Ommer.
\newblock High-resolution image synthesis with latent diffusion models.
\newblock In \emph{Proceedings of the IEEE/CVF conference on computer vision and pattern recognition}, pp.\  10684--10695, 2022.

\bibitem[Schuhmann et~al.(2022)Schuhmann, Beaumont, Vencu, Gordon, Wightman, Cherti, Coombes, Katta, Mullis, Wortsman, et~al.]{laion}
Christoph Schuhmann, Romain Beaumont, Richard Vencu, Cade Gordon, Ross Wightman, Mehdi Cherti, Theo Coombes, Aarush Katta, Clayton Mullis, Mitchell Wortsman, et~al.
\newblock Laion-5b: An open large-scale dataset for training next generation image-text models.
\newblock \emph{Advances in Neural Information Processing Systems}, 35:\penalty0 25278--25294, 2022.

\bibitem[Sidorov et~al.(2020)Sidorov, Hu, Rohrbach, and Singh]{textcaps}
Oleksii Sidorov, Ronghang Hu, Marcus Rohrbach, and Amanpreet Singh.
\newblock Textcaps: a dataset for image captioning with reading comprehension.
\newblock In \emph{Computer Vision--ECCV 2020: 16th European Conference, Glasgow, UK, August 23--28, 2020, Proceedings, Part II 16}, pp.\  742--758. Springer, 2020.

\bibitem[Song \& Ermon(2019)Song and Ermon]{song2019generative}
Yang Song and Stefano Ermon.
\newblock Generative modeling by estimating gradients of the data distribution.
\newblock \emph{Advances in neural information processing systems}, 32, 2019.

\bibitem[Sun et~al.(2024{\natexlab{a}})Sun, Jiang, Chen, Zhang, Peng, Luo, and Yuan]{llamagen}
Peize Sun, Yi~Jiang, Shoufa Chen, Shilong Zhang, Bingyue Peng, Ping Luo, and Zehuan Yuan.
\newblock Autoregressive model beats diffusion: Llama for scalable image generation.
\newblock \emph{arXiv preprint arXiv:2406.06525}, 2024{\natexlab{a}}.

\bibitem[Sun et~al.(2023)Sun, Yu, Cui, Zhang, Zhang, Wang, Gao, Liu, Huang, and Wang]{sun2023generative}
Quan Sun, Qiying Yu, Yufeng Cui, Fan Zhang, Xiaosong Zhang, Yueze Wang, Hongcheng Gao, Jingjing Liu, Tiejun Huang, and Xinlong Wang.
\newblock Generative pretraining in multimodality.
\newblock \emph{arXiv preprint arXiv:2307.05222}, 2023.

\bibitem[Sun et~al.(2024{\natexlab{b}})Sun, Bao, Wang, Peng, Dong, Huang, Wang, and Wei]{sun2024multimodal}
Yutao Sun, Hangbo Bao, Wenhui Wang, Zhiliang Peng, Li~Dong, Shaohan Huang, Jianyong Wang, and Furu Wei.
\newblock Multimodal latent language modeling with next-token diffusion.
\newblock \emph{arXiv preprint arXiv:2412.08635}, 2024{\natexlab{b}}.

\bibitem[Team(2024)]{team2024chameleon}
Chameleon Team.
\newblock Chameleon: Mixed-modal early-fusion foundation models.
\newblock \emph{arXiv preprint arXiv:2405.09818}, 2024.

\bibitem[Team et~al.(2023)Team, Anil, Borgeaud, Wu, Alayrac, Yu, Soricut, Schalkwyk, Dai, Hauth, et~al.]{team2023gemini}
Gemini Team, Rohan Anil, Sebastian Borgeaud, Yonghui Wu, Jean-Baptiste Alayrac, Jiahui Yu, Radu Soricut, Johan Schalkwyk, Andrew~M Dai, Anja Hauth, et~al.
\newblock Gemini: a family of highly capable multimodal models.
\newblock \emph{arXiv preprint arXiv:2312.11805}, 2023.

\bibitem[Tian et~al.(2024)Tian, Jiang, Yuan, Peng, and Wang]{tian2024visual}
Keyu Tian, Yi~Jiang, Zehuan Yuan, Bingyue Peng, and Liwei Wang.
\newblock Visual autoregressive modeling: Scalable image generation via next-scale prediction.
\newblock \emph{arXiv preprint arXiv:2404.02905}, 2024.

\bibitem[Tuo et~al.(2023)Tuo, Xiang, He, Geng, and Xie]{anytext}
Yuxiang Tuo, Wangmeng Xiang, Jun-Yan He, Yifeng Geng, and Xuansong Xie.
\newblock Anytext: Multilingual visual text generation and editing.
\newblock \emph{arXiv preprint arXiv:2311.03054}, 2023.

\bibitem[Wang et~al.(2025)Wang, Mao, Zhang, Han, Dong, Li, Lin, Yang, Qin, Zhang, et~al.]{wang2025textatlas5m}
Alex~Jinpeng Wang, Dongxing Mao, Jiawei Zhang, Weiming Han, Zhuobai Dong, Linjie Li, Yiqi Lin, Zhengyuan Yang, Libo Qin, Fuwei Zhang, et~al.
\newblock Textatlas5m: A large-scale dataset for dense text image generation.
\newblock \emph{arXiv preprint arXiv:2502.07870}, 2025.

\bibitem[Wang et~al.(2024)Wang, Zhang, Luo, Sun, Cui, Wang, Zhang, Wang, Li, Yu, et~al.]{wang2024emu3}
Xinlong Wang, Xiaosong Zhang, Zhengxiong Luo, Quan Sun, Yufeng Cui, Jinsheng Wang, Fan Zhang, Yueze Wang, Zhen Li, Qiying Yu, et~al.
\newblock Emu3: Next-token prediction is all you need.
\newblock \emph{arXiv preprint arXiv:2409.18869}, 2024.

\bibitem[Wu et~al.(2025)Wu, Chen, Wu, Ma, Liu, Pan, Liu, Xie, Yu, Ruan, et~al.]{wu2025janus}
Chengyue Wu, Xiaokang Chen, Zhiyu Wu, Yiyang Ma, Xingchao Liu, Zizheng Pan, Wen Liu, Zhenda Xie, Xingkai Yu, Chong Ruan, et~al.
\newblock Janus: Decoupling visual encoding for unified multimodal understanding and generation.
\newblock In \emph{Proceedings of the Computer Vision and Pattern Recognition Conference}, pp.\  12966--12977, 2025.

\bibitem[Wu et~al.(2024)Wu, Fei, Qu, Ji, and Chua]{wu2024next}
Shengqiong Wu, Hao Fei, Leigang Qu, Wei Ji, and Tat-Seng Chua.
\newblock Next-gpt: Any-to-any multimodal llm.
\newblock In \emph{Forty-first International Conference on Machine Learning}, 2024.

\bibitem[Xie et~al.(2024)Xie, Mao, Bai, Zhang, Wang, Lin, Gu, Chen, Yang, and Shou]{xie2024show}
Jinheng Xie, Weijia Mao, Zechen Bai, David~Junhao Zhang, Weihao Wang, Kevin~Qinghong Lin, Yuchao Gu, Zhijie Chen, Zhenheng Yang, and Mike~Zheng Shou.
\newblock Show-o: One single transformer to unify multimodal understanding and generation.
\newblock \emph{arXiv preprint arXiv:2408.12528}, 2024.

\bibitem[Yan et~al.(2025)Yan, Ye, Li, Huang, Yuan, He, Lin, He, He, and Yuan]{4oimagepowerful2}
Zhiyuan Yan, Junyan Ye, Weijia Li, Zilong Huang, Shenghai Yuan, Xiangyang He, Kaiqing Lin, Jun He, Conghui He, and Li~Yuan.
\newblock Gpt-imgeval: A comprehensive benchmark for diagnosing gpt4o in image generation.
\newblock \emph{arXiv preprint arXiv:2504.02782}, 2025.

\bibitem[Ye et~al.(2025)Ye, He, Li, Lin, Yuan, Yan, Hou, and Yuan]{ye2025imgedit}
Yang Ye, Xianyi He, Zongjian Li, Bin Lin, Shenghai Yuan, Zhiyuan Yan, Bohan Hou, and Li~Yuan.
\newblock Imgedit: A unified image editing dataset and benchmark.
\newblock \emph{arXiv preprint arXiv:2505.20275}, 2025.

\bibitem[Yu et~al.(2022)Yu, Xu, Koh, Luong, Baid, Wang, Vasudevan, Ku, Yang, Ayan, et~al.]{parti}
Jiahui Yu, Yuanzhong Xu, Jing~Yu Koh, Thang Luong, Gunjan Baid, Zirui Wang, Vijay Vasudevan, Alexander Ku, Yinfei Yang, Burcu~Karagol Ayan, et~al.
\newblock Scaling autoregressive models for content-rich text-to-image generation.
\newblock \emph{arXiv preprint arXiv:2206.10789}, 2\penalty0 (3):\penalty0 5, 2022.

\bibitem[Yu et~al.(2024)Yu, Chow, Yue, Pan, Wu, Wan, Li, Tang, Zhang, and Zhuang]{anyedit}
Qifan Yu, Wei Chow, Zhongqi Yue, Kaihang Pan, Yang Wu, Xiaoyang Wan, Juncheng Li, Siliang Tang, Hanwang Zhang, and Yueting Zhuang.
\newblock Anyedit: Mastering unified high-quality image editing for any idea.
\newblock \emph{arXiv preprint arXiv:2411.15738}, 2024.

\bibitem[Zhan et~al.(2024)Zhan, Dai, Ye, Zhou, Zhang, Liu, Zhang, Yuan, Zhang, Li, et~al.]{anygpt}
Jun Zhan, Junqi Dai, Jiasheng Ye, Yunhua Zhou, Dong Zhang, Zhigeng Liu, Xin Zhang, Ruibin Yuan, Ge~Zhang, Linyang Li, et~al.
\newblock Anygpt: Unified multimodal llm with discrete sequence modeling.
\newblock \emph{arXiv preprint arXiv:2402.12226}, 2024.

\bibitem[Zhang et~al.(2025)Zhang, Duan, Wang, Zhao, Lu, Di, Xu, Chen, and Zhang]{zhang2025nexusgen}
Hong Zhang, Zhongjie Duan, Xingjun Wang, Yuze Zhao, Weiyi Lu, Zhipeng Di, Yixuan Xu, Yingda Chen, and Yu~Zhang.
\newblock Nexus-gen: A unified model for image understanding, generation, and editing.
\newblock \emph{arXiv preprint arXiv:2504.21356}, 2025.

\bibitem[Zhang et~al.(2023)Zhang, Mo, Chen, Sun, and Su]{magicbrush}
Kai Zhang, Lingbo Mo, Wenhu Chen, Huan Sun, and Yu~Su.
\newblock Magicbrush: A manually annotated dataset for instruction-guided image editing.
\newblock \emph{Advances in Neural Information Processing Systems}, 36:\penalty0 31428--31449, 2023.

\bibitem[Zhao et~al.(2024)Zhao, Ma, Chen, Si, Wu, An, Yu, Zhang, Li, and Chang]{ultraedit}
Haozhe Zhao, Xiaojian~Shawn Ma, Liang Chen, Shuzheng Si, Rujie Wu, Kaikai An, Peiyu Yu, Minjia Zhang, Qing Li, and Baobao Chang.
\newblock Ultraedit: Instruction-based fine-grained image editing at scale.
\newblock \emph{Advances in Neural Information Processing Systems}, 37:\penalty0 3058--3093, 2024.

\bibitem[Zhou et~al.(2024)Zhou, Yu, Babu, Tirumala, Yasunaga, Shamis, Kahn, Ma, Zettlemoyer, and Levy]{zhou2024transfusion}
Chunting Zhou, Lili Yu, Arun Babu, Kushal Tirumala, Michihiro Yasunaga, Leonid Shamis, Jacob Kahn, Xuezhe Ma, Luke Zettlemoyer, and Omer Levy.
\newblock Transfusion: Predict the next token and diffuse images with one multi-modal model.
\newblock \emph{arXiv preprint arXiv:2408.11039}, 2024.

\bibitem[Zhuo et~al.(2024)Zhuo, Du, Xiao, Li, Liu, Huang, Liu, Zhao, Wang, Ma, et~al.]{2024lumina}
Le~Zhuo, Ruoyi Du, Han Xiao, Yangguang Li, Dongyang Liu, Rongjie Huang, Wenze Liu, Lirui Zhao, Fu-Yun Wang, Zhanyu Ma, et~al.
\newblock {Lumina-Next}: Making {Lumina-T2X} stronger and faster with {Next-DiT}.
\newblock \emph{arXiv preprint arXiv:2406.18583}, 2024.

\end{thebibliography}
\bibliographystyle{colm2025_conference}

\clearpage
\appendix


\section{Related Work}
\subsection{Instruction-guided Generation Models}
Instruction-guided generation has recently gained momentum, with its main approaches centered on diffusion models~\citep{song2019generative,ho2020denoising,rombach2022high,ho2022classifier,ge2024seed,zhang2025nexusgen,chen2025blip3o,cai2025hidream} and autoregressive generation models~\citep{ramesh2021zero,esser2021taming,wu2025janus,team2024chameleon,zhou2024transfusion,xie2024show,liu2024world,wang2024emu3,liu2024world}.
Diffusion models~\citep{ramesh2021zero,parti} iteratively denoise random noise to produce high-fidelity, diverse images, capturing fine details and complex structures.
Autoregressive models~\citep{unifiedio2,sun2023generative,anygpt,llamagen} discretize images into token sequences via vector quantization, enabling efficient, controllable generation and seamless integration with LLMs.
To broaden the scope of applications, several works~\citep{liu2025step1x,anyedit,ye2025imgedit,ultraedit,magicbrush,instructp2p} investigate instruction-driven image editing, allowing models to create new images by modifying inputs according to specified instructions.

\subsection{Datasets for Image Generation}
Early image-text pair datasets, such as MS-COCO~\citep{mscoco} and TextCaps~\citep{textcaps}, provided natural language captions aligned with images and have been widely utilized in \textit{text-to-image} generation tasks. To further enhance the performance and generalization capabilities of generative models, subsequent works~\citep{cc12m, laion, textdiffuser, anytext, wang2025textatlas5m} collected larger-scale image-text pairs from the internet, enriching the accompanying textual descriptions.
In addition to \textit{text-to-image} synthesis, certain datasets~\citep{instructp2p, hqedit, magicbrush, seeddataedit, ultraedit, anyedit, ye2025imgedit} focus specifically on image editing tasks, wherein models are required to modify images based on natural language commands. These tasks encompass operations such as action manipulation and style transfer, which support more fine-grained visual control and enable richer forms of multimodal interaction.
However, the use of real-world images as training targets may introduce challenges, including low visual quality and inconsistency. Overall, there remains a significant lack of high-quality image generation datasets that can effectively distill the capabilities of the most advanced image generation models.

\subsection{Multimodal Large Language Models (MLLMs)}
To enhance interactivity, Large Language Models (LLMs) have evolved beyond text, embracing a multimodal paradigm. With the emergence of GPT-4o~\citep{achiam2023gpt}, researchers began enabling LLMs to understand images~\citep{liu2023visual,chen2024towards,qwen2.5vl,team2023gemini,chen2024internvl} by aligning visual features with text using paired image-text data and pre-trained image encoders. Building on this, image generation became a natural extension. Early works like Vitron~\citep{fei2024vitron} treated it as tool use, where LLMs generated prompts for external models. Later, some approaches~\citep{sun2024multimodal, wu2024next, anygpt} enabled external decoders to generate images directly from LLM representations. Now, models like Janus support fully autoregressive image generation via image tokens~\cite{henighan2020scaling, tian2024visual, wu2025janus}. These models offer unique advantages: (1) a simple end-to-end design with no extra components, and (2) native LLM understanding applied to image generation, enabling better instruction following. Janus-4o will be the first of its kind to support both \textit{text-to-image} and \textit{text-and-image-to-image} generation.


\section{Image Generation Categories}
\label{app:category}
\subsection{Categories}
\subsubsection{\textit{Text-to-Image} Dimensions}
\label{app:image_text_task_categories}

\begin{table}[htbp]
    \centering
    \begin{tabular}{ll}
\toprule
\textbf{Dimension} & \textbf{Sub-dimension} \\
\midrule

\textbf{Background categories} &
  \begin{tabular}[t]{@{}l@{}} 
  a. abstract and surreal landscapes \\
  b. urban and architectural settings \\
  c. interior spaces
  \end{tabular} \\
\midrule

\textbf{Style categories} &
  \begin{tabular}[t]{@{}l@{}}
  a. cultural and historical aesthetics \\
  b. artistic movements and periods \\
  c. digital and graphic art styles
  \end{tabular} \\
\midrule

\textbf{Lighting options} &
  \begin{tabular}[t]{@{}l@{}}
  a. color temperature \\
  b. light direction \\
  c. light intensity and quality
  \end{tabular} \\
\midrule

\textbf{Camera viewpoints} &
  \begin{tabular}[t]{@{}l@{}}
  a. vertical perspective \\
  b. horizontal perspective \\
  c. shot distance and framing
  \end{tabular} \\
\midrule

\textbf{Composition techniques} &
  \begin{tabular}[t]{@{}l@{}}
  a. basic division and placement \\
  b. balance and symmetry \\
  c. framing depth and layers
  \end{tabular} \\

\bottomrule
\end{tabular}
    \caption{Text-to-Image Dimensions and Sub-dimensions.}
    \label{tab:t2i_category}
\end{table}

\begin{figure}[htbp]
    \centering
    \includegraphics[width=0.95\linewidth]{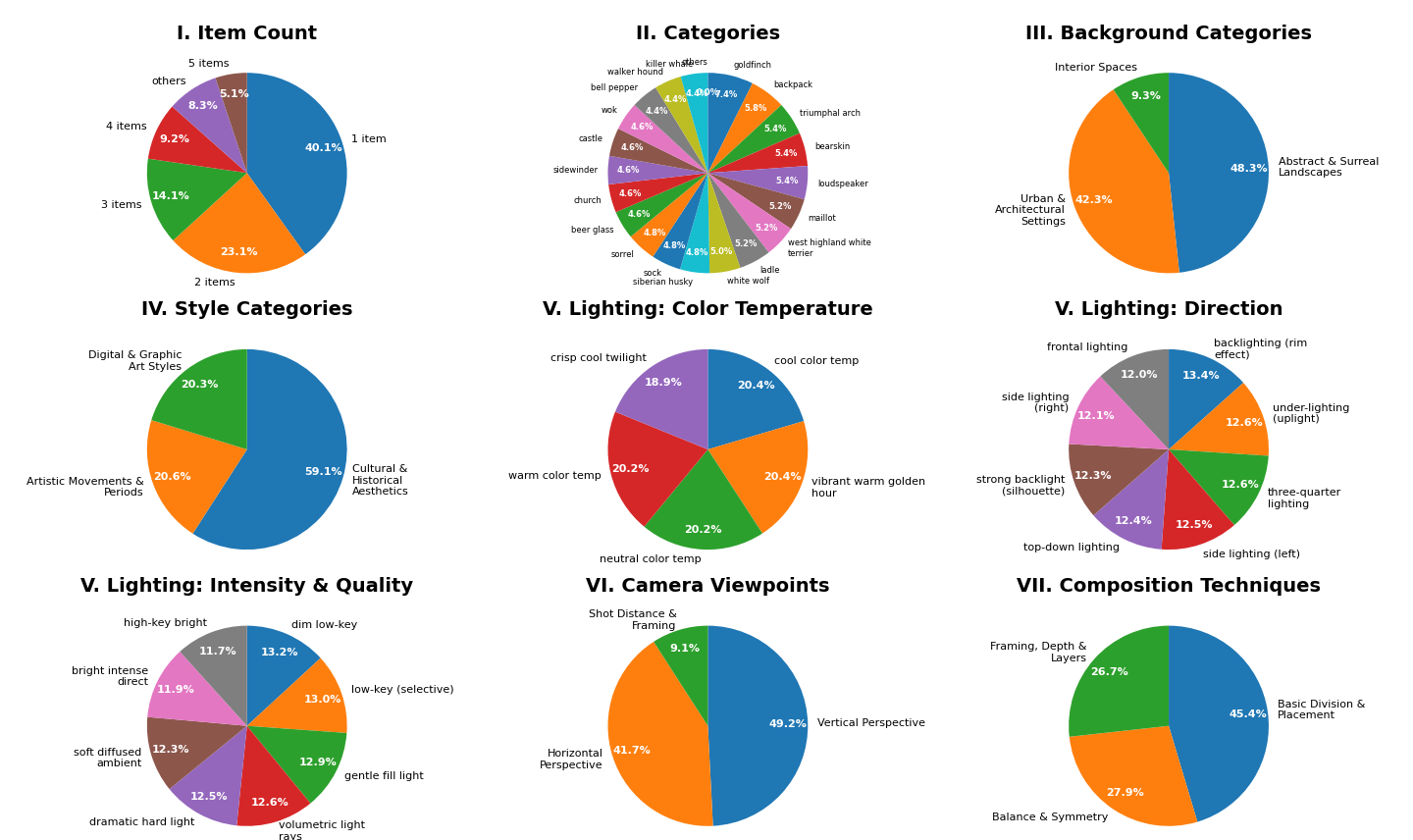}
    \caption{Text-to-Image Categories distributions.}
    \label{fig:t2i_category_distribution}
\end{figure}

Detailed Text-to-Image dimensions are shown in Table~\ref{tab:t2i_category} (apart from \textbf{Object} dimension, whose sub-dimensions can be found in ImageNet~\cite{deng2009imagenet}). The distribution among different categories can be found in Figure~\ref{fig:t2i_category_distribution}.

\subsubsection{\textit{Text-and-Image-to-Image} Categories}

\begin{table}[htbp]
\centering
\scriptsize
\begin{tabular}{ll}
\toprule
\textbf{Category} & \textbf{Subcategory} \\
\midrule
\textbf{I. Image editing and manipulation} &
  \begin{tabular}[t]{@{}l@{}}
  a. Inpainting and replacement \\
  b. Element manipulation \\
  c. Background modification \\
  d. Attribute and effect manipulation
  \end{tabular} \\
\midrule
\textbf{II. Style transfer and artistic transformation} &
  \begin{tabular}[t]{@{}l@{}}
  a. Specific artist and art styles \\
  b. Medium and technique styles \\
  c. Genre theme and era shifting
  \end{tabular} \\
\midrule
\textbf{III. Content augmentation and extension} &
  \begin{tabular}[t]{@{}l@{}}
  a. Resolution detail and quality enhancement \\
  b. Image outpainting and inpainting for extension
  \end{tabular} \\
\midrule
\textbf{IV. Struc
tured generation and conditional control} &
  \begin{tabular}[t]{@{}l@{}}
  a. From sketch lineart and edges \\
  b. From pose depth and segmentation
  \end{tabular} \\
\midrule
\textbf{V. Creative ideation and iteration} &
  \begin{tabular}[t]{@{}l@{}}
  a. Storyboarding and sequential generation \\
  b. Concept variation and exploration \\
  c. Hybridization and mashups
  \end{tabular} \\
\bottomrule
\end{tabular}
\caption{\label{tab:ti2i_categories}Categories and subcategories for \textit{Text-and-Image-to-Image} tasks.}
\end{table}

\begin{figure}
    \centering
    \includegraphics[width=0.8\linewidth]{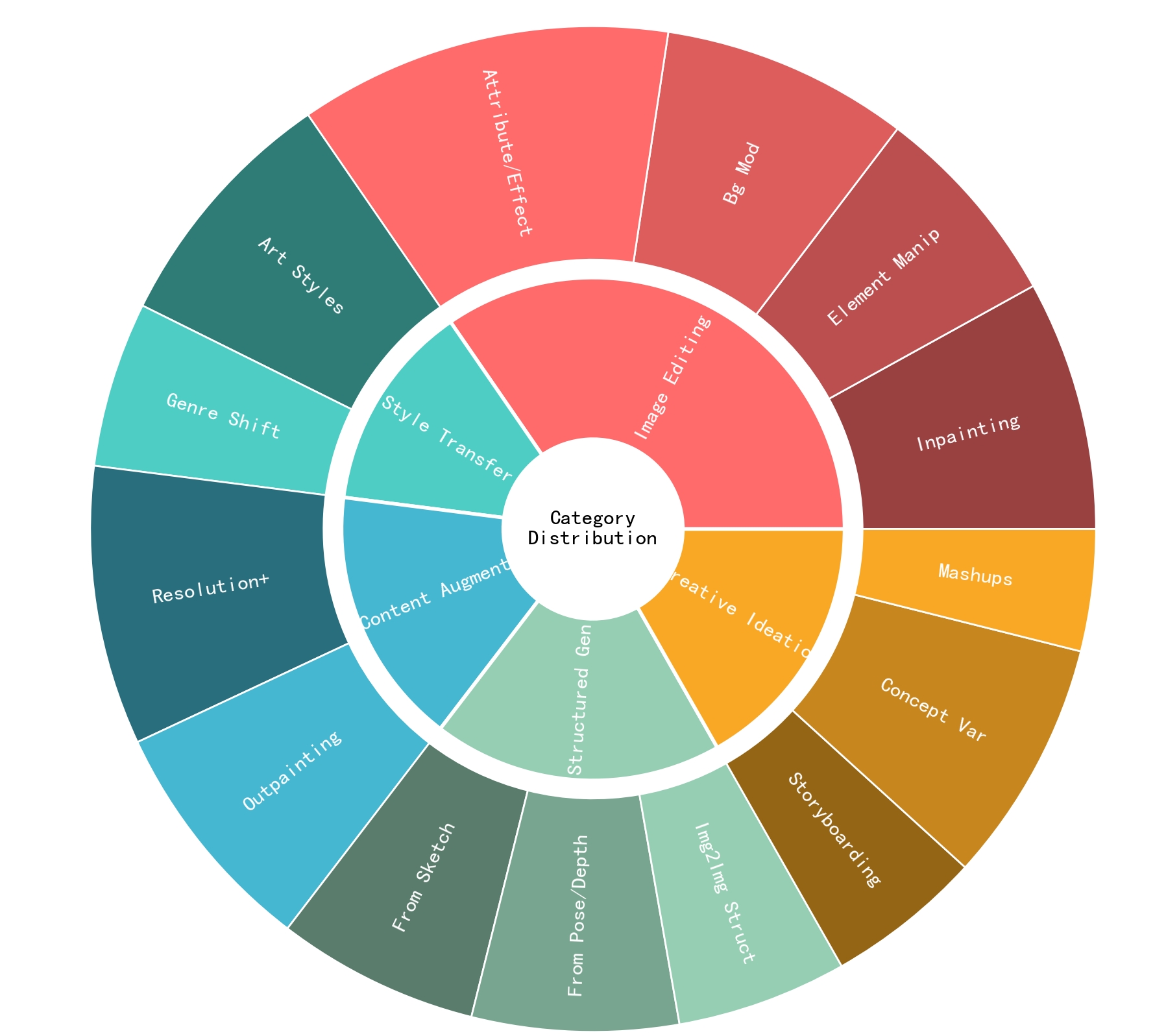}
    \caption{\texorpdfstring{Distribution of \textit{Text-and-Image-to-Image} categories.}{Distribution of Text-and-Image-to-Image categories.}}
    \label{fig:it2i_categories}
\end{figure}

Detailed categories and sub-categories for Text-and-Image-to-Image tasks can be found in Table~\ref{tab:ti2i_categories}. The distribution of the categories can be found in Figure~\ref{fig:it2i_categories}.

\subsection{\texorpdfstring{Distribution of $k$}{Distribution of k}}

In this appendix, we explain the distribution used for sampling the number of object categories, $k$. The selection is based on a random sample ranging from 1 to 100, with weights applied to ensure diverse sampling patterns, especially when multiple objects interact.

\subsubsection{Exponential Decay Distribution}

We employ an exponential decay distribution to model the probability of selecting a higher number of categories. This approach ensures that the probability decreases exponentially as the number increases, making it suitable for scenarios where fewer categories are more common.

The distribution is defined as follows:

\[ w(x) = \exp(-\lambda(x - 1)) \]

Where:
- \( \lambda \) is the parameter controlling the decay rate. A larger \( \lambda \) results in faster decay.

The following figure illustrates the characteristics of this distribution:

\begin{figure}[ht]
    \centering
    \includegraphics[width=\linewidth]{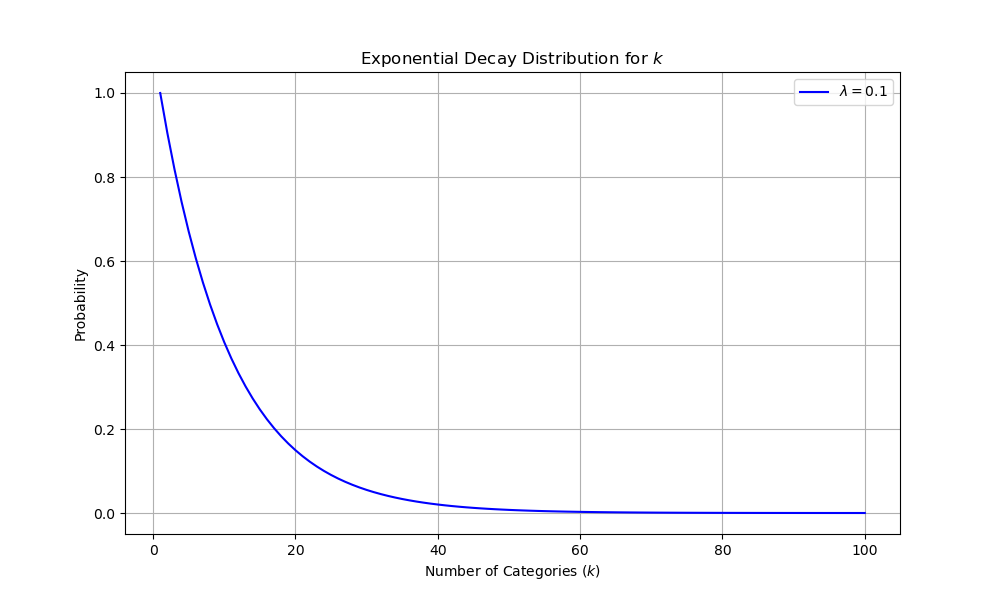} 
    \caption{Illustration of the exponential decay distribution used for sampling $k$.}
    \label{fig:exp_decay_distribution}
\end{figure}

\section{Prompts for Generation}
\label{app:generation_prompts}
\subsection{Image-First Prompt}
\label{app:image_first_prompt}
\begin{lstlisting}[language=Python]
Please describe the main content of the image in one sentence. This sentence will be used as a prompt to regenerate the image, so it should clearly capture the key visual information. Only provide the sentence,no extra text.
\end{lstlisting}

\subsection{Content Generation Process}
\label{app:document_prompt}

\begin{lstlisting}[language=Python]
FREESTYLE_VISUAL_THEME_AND_TYPE_META_PROMPT = f"""
You are an expert document concept designer and text-to-image prompt generator. Your goal is to create concepts that are clear, professional, and highly appropriate for the intended use.

Your task is to perform the following steps IN ORDER:

STEP 1: Use the Given Theme and Choose a Core Document Category
   - Use the following theme: "{{selected_theme}}"
   - Choose ONE core document category that best fits this theme from the following list:
     {', '.join(CORE_DOCUMENT_CATEGORIES)}
   - Based on the chosen core category and theme, **conceive and clearly state a MORE SPECIFIC document type or application.**
     For example:
       [previous examples remain the same...]

STEP 2: Select an Appropriate Font Style and Visual Characteristics
   - Choose and incorporate this font style: "{{selected_font}}"
   - Based on the **Specific Document Type** from STEP 1 and the Theme, internally determine the most FITTING and EFFECTIVE visual style characteristics.
   - **Priority is on CLARITY, PROFESSIONALISM, and FUNCTIONALITY for the specific document type.**
   - Avoid overly ornate or unnecessarily complex styles unless the theme and specific type genuinely call for it (e.g., a historical document recreation).
   - **Minimize generic "parchment paper" or "old paper" styles for most modern professional contexts.**
   - The visual style should serve the content and the document's purpose.

STEP 3: Generate a Clear and Fitting Visual Template Prompt
   - Now, generate a single, compelling text-to-image prompt (in English) for creating a visual template.
   - This prompt MUST be based on the Theme AND the **Specific Document Type** (from STEP 1).
   - The prompt MUST reflect a visual style that is **highly appropriate, clear, and professional** for the specific document type and theme (informed by STEP 2).
   - The prompt should describe:
     - Overall aesthetic (e.g., "clean and modern," "formal and academic," "bold and clear," "data-focused and organized").
     - Layout and composition (e.g., "standard slide layout with title and content areas," "two-column academic paper layout," "grid-based infographic structure").
     - Color palette (e.g., "corporate blues and grays," "high-contrast for readability," "theme-appropriate accent colors").
     - Textures (if any: "smooth untextured background," "subtle professional texture like fine linen," "no distracting textures").
     - Placeholder elements and their style, suitable for the specific document type (e.g., "placeholders for a large title, speaker name, and event logo for a title slide").
   - Do NOT include any specific readable text in this visual template prompt itself.
   - The prompt should be a single, coherent sentence.

STEP 4: State Your Outputs Clearly
   - Provide your outputs in the following exact format, with each item on a new line:
     Visual Template Prompt: [The visual template prompt you generated in STEP 3]
     Conceived Theme: [The theme you conceived in STEP 1]
     Document Type: [The CORE DOCUMENT CATEGORY chosen in STEP 1, followed by the SPECIFIC document type/application you conceived in STEP 1, e.g., "Slide - Title Slide for a Tech Conference"]

Now, follow these steps carefully, focusing on CLARITY, PROFESSIONALISM, and APPROPRIATENESS for the document type, and generate a new set:
"""
\end{lstlisting}

\subsection{Meta Prompts for Image Text Instruction}
\label{app:image_text_instruction_meta_prompts}

\begin{lstlisting}[language=Python]
meta_prompt_system = (
    "You are an AI assistant that creates a single, concise, complete, and descriptive English sentence to be used as a prompt for image generation models. "
    "This sentence should beautifully capture the essence, mood, and key visual elements of the provided input image. "
    "If a 'Base Task' is provided, consider integrating it naturally into this single descriptive sentence. "
    "However, prioritizing the image's core visual appeal and the formation of a well-structured, impactful sentence is more important than rigidly following the task. "
    "If the task doesn't fit well or makes the sentence awkward or verbose, focus on crafting the best single descriptive sentence for the image itself. "
    "Output ONLY this single English sentence. No explanations, no preambles, no bullet points, no numbered lists. Just one sentence. "
    "For example, for an image of a misty forest (task: 'add creature'): 'A mystical deer steps softly through the sun-dappled, misty ancient forest.' "
    "Or if the task is awkward: 'Sunlight filters mysteriously through the tall trees in the quiet, misty forest.' "
    "For a portrait (task: 'change background to space'): 'Her serene gaze is complemented by a backdrop of swirling cosmic stardust.' "
    "Or without task: 'A soft light illuminates her thoughtful expression in this intimate portrait.'"
)
\end{lstlisting}

\section{Document Pipeline}
\label{app:document_pipeline}

\subsection{Pipeline}
\paragraph{Document Prompts Generation.}
To generate realistic document-style images (e.g., slides, posters, reports), we implemented a specialized workflow:
\begin{enumerate}
    \item \textbf{Content Generation}: The LLM first generates plausible textual and structural content for the document (e.g., titles, paragraphs, data snippets). This is guided by meta-prompts tailored to different document types (see Appendix~\ref{app:document_type} for examples).
    \item \textbf{Prompt Synthesis}: Based on this generated content, the LLM then formulates a detailed textual prompt describing the desired visual layout, typography, color scheme, and overall appearance of the document image. The meta-prompts used for this stage are also provided in Appendix~\ref{app:document_prompt}.
\end{enumerate}
This specialized approach helps the prompt for document images to capture the unique compositional and semantic nuances inherent to this image category.

\subsection{Core Document Categories}
\label{app:document_type}

\begin{itemize}
    \item Slide
    \item Paper
    \item Document
    \item Infographic/Chart
    \item Poster
\end{itemize}


\section{Ethical Considerations and Societal Impact}

The development and release of ShareGPT-4o-Image are intended to accelerate open-source research in multimodal AI. However, we acknowledge several ethical considerations pertinent to its construction and potential use.

\textbf{Data Source and Transparency:} The image data in ShareGPT-4o-Image are synthesized outputs from GPT-4o-Image, a proprietary model. While our methodology for prompt generation is detailed extensively (Section 2.1, Appendices \ref{app:category}-\ref{app:generation_prompts}) to ensure transparency and facilitate reproducibility of the *type* of data generated, the internal workings of GPT-4o-Image remain opaque. Our dataset aims to transfer capabilities, not to replicate the proprietary model itself.

\textbf{Potential for Bias in Generated Images:} Large-scale generative models, including GPT-4o-Image, are trained on vast web-scale datasets and may inherit or amplify existing societal biases related to gender, race, age, occupation, and other demographic attributes. Consequently, images generated by GPT-4o-Image, particularly those depicting human subjects, may reflect such biases.
\begin{itemize}
    \item \textbf{Mitigation Efforts during Prompting:} While curating attributes for the "Person" category (Appendix~\ref{app:category}), we aimed for a diverse set of descriptors for aspects like ethnicity and age. However, the efficacy of these attributes in ensuring perfectly balanced representation in the generated images is subject to the behavior of the underlying GPT-4o-Image model.
    \item \textbf{No Post-hoc Filtering for Bias:} We have not performed post-hoc filtering of images based on perceived bias, as this would involve subjective judgments and could itself introduce new biases. Instead, we advocate for awareness and downstream bias analysis by users of the dataset.
\end{itemize}

\textbf{Content Generation:} The textual content for document prompts and the synthesis of descriptive prompts were performed by an LLM (Gemini 2.5 Pro). While prompts were designed to generate generic and fictional content, there is a remote possibility of generating text that could be misconstrued or reflect unintentional biases from the LLM's training data.

\textbf{Intended Use and Misuse Potential:} ShareGPT-4o-Image is released for academic research purposes to study instruction following, image quality, and multimodal conditioning in open-source generative models. We explicitly discourage any use of this dataset or models trained on it for creating harmful content, perpetuating stereotypes, generating misinformation, or infringing on the rights of individuals. Users of the dataset are expected to adhere to ethical AI practices.

\textbf{Future Work and Community Responsibility:} We encourage further research into the characteristics of ShareGPT-4o-Image, including systematic bias audits. The release of this dataset aims to empower the open-source community, and with this empowerment comes a shared responsibility to investigate and mitigate potential negative societal impacts of advanced generative AI.

\end{document}